\documentclass[journal]{IEEEtran}

\usepackage{framed,multirow}
\usepackage{times}
\usepackage{epsfig}
\usepackage{graphicx}
\usepackage{amsmath}
\usepackage{amssymb}
\usepackage{caption}
\usepackage{float}
\usepackage{placeins}
\usepackage{xspace}
\usepackage{multirow}
\usepackage{latexsym}

\usepackage{pifont}
\usepackage{amsmath}
\usepackage{amssymb}
\usepackage{color,soul}

\newcommand{\mytilde}{\raise.17ex\hbox{$\scriptstyle\mathtt{\sim}$}}

\def\ee{\mathbf{e}}
\def\ff{\mathbf{f}}

\def\jj{\mathbf{j}}

\def\ll{\mathbf{l}}

\def\nn{\mathbf{n}}

\def\pp{\mathbf{p}}

\def\sss{\mathbf{s}}
\def\ttt{\mathbf{t}}

\def\vv{\mathbf{v}}

\def\AA{\mathbf{A}}

\def\CC{\mathbf{C}}
\def\DD{\mathbf{D}}
\def\EE{\mathbf{E}}

\def\II{\mathbf{I}}
\def\JJ{\mathbf{J}}
\def\KK{\mathbf{K}}
\def\LL{\mathbf{L}}
\def\MM{\mathbf{M}}

\def\OO{\mathbf{O}}
\def\PP{\mathbf{P}}
\def\QQ{\mathbf{Q}}

\def\SS{\mathbf{S}}

\def\VV{\mathbf{V}}
\def\WW{\mathbf{W}}

\def\cC{\mathcal{C}}

\def\Re{\mathbb{R}}

\def\modelname{MeT\xspace}

\begin{document}

\title{\textit{\modelname}: A Graph Transformer for Semantic Segmentation of 3D Meshes}

\author{%
  \IEEEauthorblockN{%
    Giuseppe Vecchio\IEEEauthorrefmark{1},
    Luca Prezzavento\IEEEauthorrefmark{1},
    Carmelo Pino\IEEEauthorrefmark{2},
    Francesco Rundo\IEEEauthorrefmark{2}, \\
    Simone Palazzo\IEEEauthorrefmark{1},
    Concetto Spampinato\IEEEauthorrefmark{1}
  } \\
  \IEEEauthorblockA{\IEEEauthorrefmark{1} Department of Computer Engineering, University of Catania} \\
  \IEEEauthorblockA{\IEEEauthorrefmark{2} ADG, R\&D Power and Discretes, STMicroelectronics}
}

\maketitle

\begin{abstract}
Polygonal meshes have become the standard for discretely approximating 3D shapes, thanks to their efficiency and high flexibility in capturing non-uniform shapes. This non-uniformity, however, leads to irregularity in the mesh structure, making tasks like segmentation of 3D meshes particularly challenging. 
Semantic segmentation of 3D mesh has been typically addressed through CNN-based approaches, leading to good accuracy. 
Recently, transformers have gained enough momentum both in NLP and computer vision fields, achieving performance at least on par with CNN models, supporting the long-sought architecture universalism. 
Following this trend, we propose a transformer-based method for semantic segmentation of 3D mesh motivated by a better modeling of the graph structure of meshes, by means of global attention mechanisms.
In order to address the limitations of standard transformer architectures in modeling relative positions of non-sequential data, as in the case of 3D meshes, as well as in capturing the local context, we perform positional encoding by means the Laplacian eigenvectors of the adjacency matrix, replacing the traditional sinusoidal positional encodings, and by introducing clustering-based features into the self-attention and cross-attention operators.
Experimental results, carried out on three sets of the Shape COSEG Dataset~\cite{wang2012active}, on the human segmentation dataset proposed in~\cite{maron2017convolutional} and on the ShapeNet benchmark~\cite{chang2015shapenet}, show how the proposed approach yields state-of-the-art performance on semantic segmentation of 3D meshes.  
\end{abstract}

\section{Introduction}
\label{sec:introduction}

Three-dimensional (3D) shapes are at the core of computer graphics and play an important role in many daily-life applications such as vision, robotics, medicine, augmented reality, and virtual reality. 
In recent years, many approaches have been proposed to encode real-world shapes, including 3D meshes~\cite{botsch2010polygon} and point clouds~\cite{nguyen20133d}.
Meshes have become widely adopted to represent complex real-world objects, which are commonly composed of continuous surfaces, through a discrete approximation. 
The mesh is an efficient way to represent non-uniform surfaces, from simple shapes that generally require only a small number of polygons, up to arbitrarily complex objects, where the number of required polygons may increase significantly. The advantages presented by a mesh are particularly evident when compared to other forms of representation, like point clouds, which fall short when higher quality and preservation of sharp shape features are required.

With the increasing spread of deep learning techniques in many fields, research has tried to apply approaches from computer vision to 3D shape analysis. Convolutional neural networks (CNNs), in particular, have demonstrated outstanding performance on a variety of images-related tasks such as classification~\cite{krizhevsky2012imagenet,szegedy2016rethinking,he2016deep} and semantic segmentation~\cite{ronneberger2015u,jegou2017one,chen2017deeplab}. However, CNNs are designed to work on images, which are represented on a regular grid of discrete values, far from the irregular representation of 3D shapes. On the other hand, representing 3D objects through volumetric grids, e.g. mapping 3D shapes to multiple 2D
projections~\cite{su2015multi} or 3D voxel grids~\cite{wu20153d}, is extremely inefficient and leads to computational costs that increase exponentially with higher resolution.

Recent approaches have tried to directly apply CNNs to the sparse point cloud representation~\cite{qi2017pointnet, achlioptas2018learning}. These approaches have a substantial gain in terms of efficiency, but present an ill-defined notion of neighborhoods and connectivity and are inherently oblivious to the local surface. This issue makes the application of convolution and pooling operations non-trivial. 
To overcome this limitation, several works have recently tried to generalize CNN architectures to non-Euclidean domains such as graphs, and incorporate neighborhood information~\cite{monti2017geometric,wang2019dynamic,li2018pointcnn}.
Other approaches have tried to apply deep neural networks to 3D meshes~\cite{boscaini2016learning,xu2017directionally,hanocka2019meshcnn}. One recent example is MeshCNN~\cite{hanocka2019meshcnn}, which obtained state-of-the-art results on several segmentation datasets.

A recent trend in computer vision revolves around the use of transformer--based architectures, originally born for NLP,~\cite{vaswani2017attention} for vision tasks~\cite{dosovitskiy2020image,touvron2021training}. The success of transformers lies in their extensive attention mechanism, which allows the network to learn global correlations between inputs. This property makes transformers able to intrinsically operate on fully-connected graphs.
However, when dealing with sparse graphs, transformers show evident limitations, mainly because of the sinusoidal positional encoding that is not able to exploit 
graph topology and to the lack of local attention operators. Recently, \cite{dwivedi2020generalization} proposed an approach to extend the transformer architecture for arbitrary graphs. It introduces a graph transformer architecture which leverages the graph connectivity inductive bias, exploiting the graph topology. In particular, they 1) propose a new attention mechanism, 2) replace the positional encoding with the Laplacian eigenvectors, 3) re-introduce batch normalization layers, and 4) take into account edge feature representation.

Inspired by this work, and leveraging the structure of a mesh, which can be represented as a graph where the nodes correspond to vertices connected by polygon edges, we propose \modelname, a transformer-based architecture for semantic mesh segmentation. In particular, our approach embeds locality features by means of the Laplacian operator (as in~\cite{dwivedi2020generalization}) and by combining polygon features with clustering-based features into a novel two-stream transformer layer architectures, where features from the two modalities are extracted through self-attention and combined through cross-attention. Additionally, we ensure that graph structure inferred by the input mesh affects the attention operators, by injecting adjacency and clustering information as attention masks.

We evaluate our method on a variety of publicly-available mesh datasets of 3D objects and human bodies; in our experiments, the proposed approach is able to  outperform previous works, both qualitatively and quantitatively.

To sum up, the key contributions of this work are:
\begin{itemize}
\item We enforce graph locality in the transformer by a combination of clustering information operator with Laplacian positional encoding in place of positional encoding.
\item We introduce novel self-attention and cross-attention mechanisms, specifically designed for mesh segmentation, that take into account adjacency and clustering information to mask elements and further impose locality.
\item Experimental results on multiple standard benchmarks  with different type of meshes showing that our model outperforms, both quantitatively and qualitatively, existing mesh segmentation methods, setting new state-of-the-art performance on the task.
\end{itemize}

\section{Related work}
\label{sec:related}

Meshes represent a way to describe 3D objects. They consist of vertices, edges and faces that defines the shape of a polyhedral object. In this work we will focus on triangular meshes, i.e., a mesh where all the faces are triangles.

\subsection{Mesh segmentation}
The semantic segmentation of 3D meshes is the process of assigning a label to each face. The task of semantic segmentation for meshes has applications in many fields such as robotics, autonomous driving, augmented reality and medical images analysis. 
Following the success of deep learning, several CNN-based methods have been applied 3D meshes to tackle the task of mesh segmentation~\cite{rodrigues2018part, he2021deep}.
We hereby present an overview of relevant work on 3D data analysis using neural networks, grouped by input representation type.

\noindent \textbf{Volumetric.} A common approach to represent the 3D shape into a binary voxel form that is the 3D analogous to a 2D grid such as an image. This allows for extending to 3D grids operations that are applied on 2D grids, thus applying any common image-based approaches to the shape domain. This concept was first introduced by~\cite{wu20153d}, who present a CNN that processes voxelized shapes for classification and completion.
Following this approach,~\cite{brock2016generative} introduce a shape reconstruction method, using a voxel-based variational autoencoder. In 2019,~\cite{hanocka2018alignet} present Alignet which used a voxel representations estimated applied the deformation on the original mesh.

Although being easy to process and extend existing method to voxels, this kind of representation is computationally and memory expensive. Resource efficient methods to process volumetric representations are an open research field with several approaches being proposed~\cite{li2016fpnn,riegler2017octnet}.
Sparse convolutions allows to further reduce computational and memory requirements, leading to more efficient approaches~\cite{yan2018second,graham20183d,choy20194d,lang2019pointpillars,yin2021center}, but suffer from inaccurate
position information due to voxelization.

\noindent \textbf{Graph.}
Another family of approaches leverages the ability to represent meshes as a graph structure. 
We distinguish between two main approaches for graph processing, one relies on the spectral properties of graphs~\cite{bruna2013spectral,henaff2015deep,defferrard2016convolutional,kostrikov2018surface}; the second one is to directly process graphs extracting locally connected regions and transforming them into a canonical form for a neural network~\cite{niepert2016learning}.
In 2017,~\cite{xu2017directionally} propose a new architecture called Directionally Convolutional Network (DCN) that extends CNNs by introducing a rotation-invariant convolution and a pooling operation on the surface of 3D shapes. In particular, they propose a two-stream segmentation framework: one stream uses the proposed DCN with face normals as the input, while the other one is implemented by a neural network operating on the face distance histogram. The learned shape representations from the two streams are fused by an element-wise product. Finally, Conditional Random Field (CRF) is applied to optimize the segmentation.
\cite{yi2017syncspeccnn} propose SyncSpecCNN, a spectral CNN with weight sharing in the spectral domain spanned by graph laplacian eigenbases, to tackle the task of 3D segmentation. 
\cite{kostrikov2018surface} propose a Graph Neural Network (GNN) which exploits the Dirac operator to leverage extrinsic differential geometry properties of three-dimensional surfaces. These methods generally operate on the vertices of a graph.

\noindent \textbf{Manifold.}
\cite{masci2015shapenet}, with the Geodesic Convolutional Neural Networks, and~\cite{boscaini2016learning} with the Anisotropic Convolutional Neural Networks, proposed two different CNNs--based architectures for triangular mesh segmentation. \\
In 2019, MeshNet was proposed by \cite{hanocka2019meshcnn}. This architecture differs from the previous by working on mesh edges rather than faces. MeshCNN combines specialized convolution and pooling layers that operate on the mesh edges by leveraging their intrinsic geodesic connections. Convolutions are applied on edges and the four edges of their incident triangles, and pooling is applied via an edge collapse operation that retains surface topology, thereby, generating new mesh connectivity for the subsequent convolutions. MeshCNN learns which edges to collapse, thus forming a task-driven process where the network exposes and expands the important features while discarding the redundant ones.

Other approaches, like~\cite{meshwalker, smirnov2021hodgenet, sharp2022diffusionnet} propose alternative solutions to the segmentation task. MeshWalker~\cite{meshwalker} represents mesh's geometry and topology by a set of random walks along the surface; these walks are fed to a recurrent neural network. HodgeNet~\cite{smirnov2021hodgenet}, instead, tackles the problem relying on spectral geometry, and proposes parallelizable algorithms for differentiating eigencomputation, including approximate backpropagation without sparse computation. Finally, DiffusionNet~\cite{sharp2022diffusionnet} introduces a general-purpose approach to deep learning on 3D surfaces, using a simple diffusion layer to agnostically represent any mesh.

\subsection{Graph transformers}
Since their introduction, Transformers~\cite{vaswani2017attention} have demonstrated their wide applicability to many different tasks, from NLP to Computer Vision. 
The original transformer was designed for handling sequential data in NLP, and operates on fully connected graphs representing all connections between the words in a sentence. However, when dealing with sparse graph, transformers perform poorly.
Recently, several attempts to adapt transformers to graphs have been proposed \cite{li2018graph,nguyen2019universal,zhang2020graph} focusing on heterogeneous graphs, temporal networks and generative modeling~\cite{yun2019graph,xu2021multigraph,hu2020heterogeneous,zhou2020data}.
In 2019, \cite{li2018graph} introduce a model employing attention an all graph nodes, instead of a node’s local neighbors, to capture global information. This approach limits the exploitation of sparsity, which is a good inductive bias for learning on graph datasets as shown in~\cite{dwivedi2020generalization}.
To learn global information other approaches involve the use of a graph-specific positional features~\cite{zhang2020graph}, node Laplacian position eigenvectors~\cite{belkin2003laplacian, dwivedi2020generalization}, relative learnable positional information~\cite{you2019position} and virtual nodes~\cite{li2015gated}.\\
\cite{dwivedi2020generalization}, propose an  approach to extend the transformer architecture for arbitrary graphs. It introduces a graph transformer architecture with four new properties compared to the standard model, which are:
1) an attention mechanism which is a function of the neighborhood connectivity for each node in the graph; 2) positional encoding represented by the Laplacian eigenvectors, which naturally generalize the sinusoidal positional encoding often used in NLP; 3) a batch normalization layer in contrast to the layer normalization; 4) edge feature representation. \\
MeshFormer~\cite{li2022meshformer} propose a mesh segmentation method based on graph transformers, which uses a boundary-preserving simplification to reduce the data size, a Ricci flow-based clustering algorithm for constructing hierarchical structures of meshes, and a graph transformer with
cross-resolution convolutions, which extracts richer high-resolution semantic.
Recently~\cite{zhuang2022ngd} introduced a novel method for 3D mesh segmentaion named Navigation Geodesic Distance Transformer (NGD-Transformer). It exploit the manifold properties of the mesh through a novel positional encoding called navigation geodesic distance positional encoding, which encodes the geodesic distance between vertices.
Our work takes inspiration from~\cite{dwivedi2020generalization} and proposes a transformer-based architecture for tackling 3D meshes represented as graphs. As in~\cite{dwivedi2020generalization} we employ a positional encoding represented by the Laplacian eigenvectors of the adjacency matrix and a pre-layer batch normalization. However, we extend the original approach by adapting the architecture to 3D meshes, particularly, by proposing two cross-attention modules (similarly to decoder layers) learning local and global representations on 3D meshes and clusters thereof.
\section{Method}
\label{sec:method}

In this work, we propose a novel transformer-based architecture for semantic segmentation of 3D meshes. The proposed method takes inspiration from recent vision transformer architectures~\cite{guo2019star} and spectral methods for graphs, in order to create an embedding in a Euclidean space with the topological features of the mesh.

Given a triangular mesh, described as a set of $V$ vertices $\left\{\vv_k = \left(x_k, y_k, z_k\right)\right\}_{k=1,\dots,V}$ and a set of $N$ triangles $\{\ff_i = (k_{i,1},k_{i,2},k_{i,3}, \nn_i)\}_{i=1,\dots,N}$, where each triangle is defined by its three vertices and the normal direction $\nn_i$ of its surface, the goal is to assign a class $c_i\in\cC$ to each triangle $\ff_i$, representing the dominant class on the surface of the triangle.

\subsection{Feature extraction}
For each triangle, we initially extract a set of features based on spectral properties of the triangle graph (where triangles are nodes, and shared sides are edges), which is the dual of the mesh (where vertices are nodes, and triangle sides are edges). 

The process starts by building the adjacency matrix $\AA$, of size $N\times N$ ($N$ being the number of triangles in the mesh), such that $A_{ij}=1$ if the $i$-th and the $j$-th triangles share an edge, and $A_{ij}=0$ otherwise. 
From the the adjacency matrix $\AA$, we then compute the symmetric normalized Laplacian matrix $\LL$ as:
\begin{equation}
    \LL = \II-\DD^{-1/2}\AA\DD^{-1/2},
\end{equation}
where $\II$ is the identity matrix and $\DD$ is the degree matrix for $\AA$, i.e., a diagonal matrix such that $D_{ii}$ is the number of edges connected to $i$ (equivalently, the sum of the elements in the $i$-th row of $\AA$).
Then, we identify the $(E+1)$-th (with $E \le N$) eigenvector with the smallest non-zero eigenvalue. The $i$-th components of the $E$ remaining eigenvectors, corresponding to vector $\ll_i$, are then used to encode the location of the $i$-th triangle within the mesh. We employ these features as a positional encoding in the transformer, as described by~\cite{dwivedi2020generalization}, and to identify local neighborhood by means of clustering (described in the next section).

Formally, given a triangle $\ff_i=(k_1,k_2,k_3,\nn_i)$, where $\nn_i$ is its normal vector direction, obtained by computing the vector product $\tilde{\nn}_i=(\vv_{k_2}-\vv_{k_1})\times(\vv_{k_3}-\vv_{k_1})$ and then normalizing it as $\nn_i = \tilde{\nn_i}/\left\| \tilde{\nn_i} \right\|$, we obtain the feature representation $\ttt_i$ for triangle $i$ as $\ttt_i = (\vv_{k_1}, \vv_{k_2}, \vv_{k_3}, \nn_i, \ll_i)$.

\begin{figure}
\begin{center}
\includegraphics[width=0.9\linewidth]{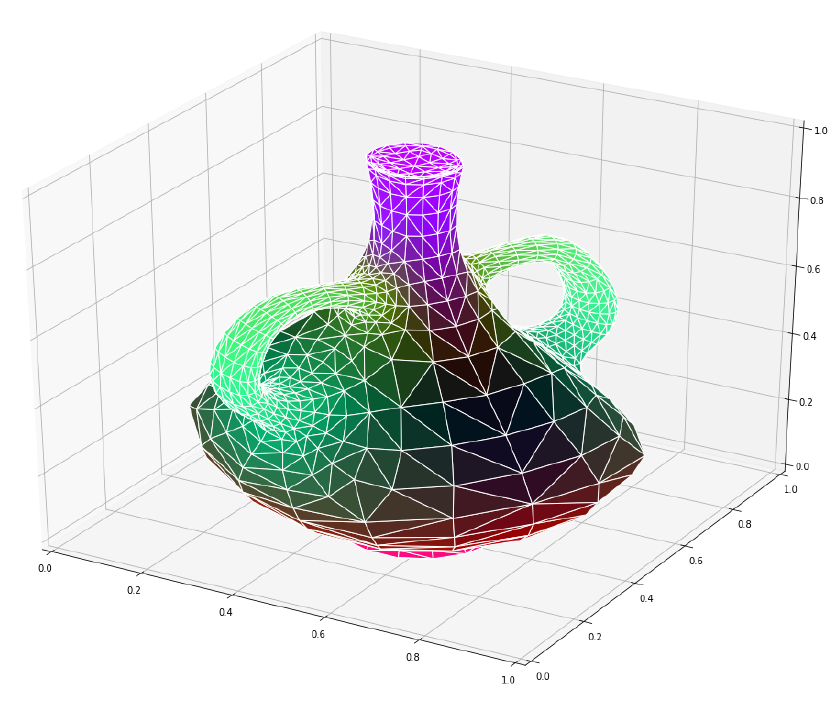}
\end{center}
   \caption{Visualization of the first three eigenvectors of the Laplacian of the mesh dual graph.}
\label{fig:laplacian}
\end{figure}

Fig. \ref{fig:laplacian} shows the visualization of the Laplacian eigenvectors for a mesh.

\subsection{Clustering}

Before being processed by the network, triangles' features are clustered in $M=V/\lambda$ clusters, where $\lambda$ is a configurable parameter, controlling the average number of mesh vertices per cluster. As we describe in detail when presenting our transformer architecture, we introduce clustering as an additional and more explicit way than positional encoding to enforce locality on the features extracted for each triangle. Clustering is carried out using the Ward method~\cite{Ward63}, which applies constraints on the connectivity dictated by the dual graph adjacency matrix, generating clusters geometrically and topologically connected and cohesive.

\begin{figure}
\begin{center}
\includegraphics[width=0.9\linewidth]{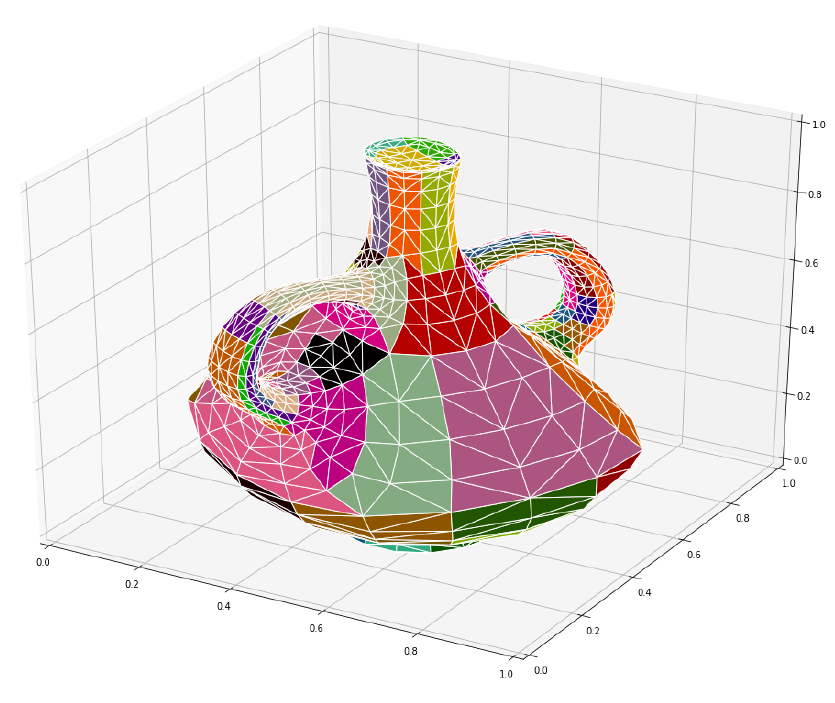}
\end{center}
   \caption{Example of triangles clustering with $\lambda = 8$.}
\label{fig:clustering}
\end{figure}

The result is a matrix $\JJ$ with shape $N\times M$, such that $J_{im} = 1$ if the $i$-th triangle is in the $m$-th cluster, and $J_{im} = 0$ otherwise. \emph{Each row $\jj_i$ in $\JJ$ can be interpreted as the one-hot cluster representation for the $i$-th triangle.}
Fig. \ref{fig:clustering} shows an example of mesh triangles clustering.

\begin{figure*}
\begin{center}
\includegraphics[width=0.8\linewidth]{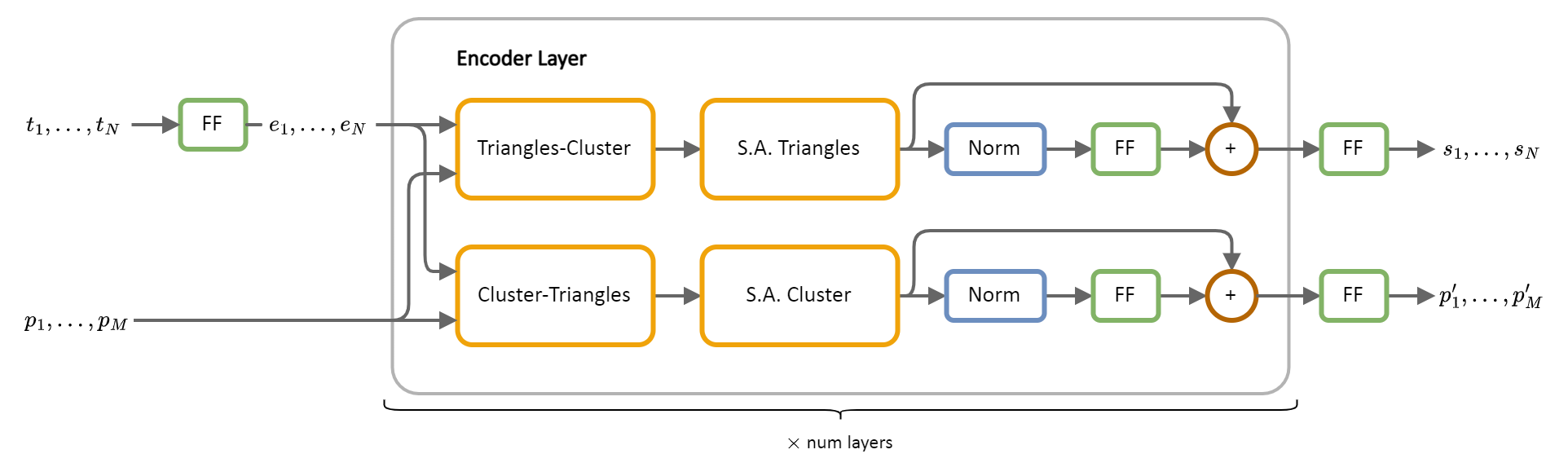}
\end{center}
   \caption{Representation of an encoder layer of the Mesh Transformer.}
\label{fig:transformer_layer}
\end{figure*}

\subsection{Network architecture}

\noindent \textbf{Mesh Transformer.} The proposed \modelname architecture implements a transformer model with two internal feature extraction streams, one for triangle features and one for cluster features, organized in matching sets (i.e., the $i$-th element of the triangle set corresponds to the $i$-th element of the cluster set). The two sets of features are processed by a cascade of transformer layers; only features from the triangle stream are finally used for prediction through a two-layer feedforward network, which predicts for each triangle a score vector $\sss_i$, of size equal to the number of segmentation classes.

Given the extracted features $\ttt_i$ and cluster identifier $\jj_i$ for each mesh triangle, we first convert them into two sequences of \emph{tokens}, to be provided as input to the transformer layers. Triangle tokens $\ee_i$ are obtained as:
\begin{equation}
\ee_i = \text{FF}_t\left( \ttt_i \right)
\end{equation}
where $\text{FF}_t$ is a feedforward layer with ReLU activation\footnote{All feedforward layers in our model have ReLU activations.}, of output size $d_t$. Cluster tokens $\pp_i$, of size $d_p$, are obtained by a learnable embedding layer on the corresponding one-hot cluster identifier $\jj_i$. Matrices $\EE \in \Re^{N \times d_t}$ and $\PP \in \Re^{N \times d_p}$ are defined by laying each token as a row in the corresponding matrix.

Each network layer, illustrated in Fig.~\ref{fig:transformer_layer}, can thus be defined as a function $L_i\left(\cdot, \cdot\right)$ on token sequences:
\begin{equation}
L_i \left( \EE, \PP \right) =
\biggl(
\text{R}_i\Bigl( \text{SA}_{t,i}\bigl( \text{TC}_i\left( \EE, \PP \right) \bigr) \Bigr),
\text{R}_i\Bigl( \text{SA}_{p,i}\bigl( \text{CT}_i\left( \EE, \PP \right) \bigr) \Bigr)
\biggr)
\end{equation}
where $\text{SA}_{t,i}$ and $\text{SA}_{p,i}$ are, respectively, the triangle and cluster self-attention functions, $\text{R}_i$ is a residual connection function, and $\text{TC}_i$ and $\text{CT}_i$ are, respectively, the function updating triangle tokens from cluster tokens and vice versa. The output of each layer has the same dimensions as the input, allowing for arbitrary length of encoder sequences.

\noindent \textbf{Multi-head attention.} Before introducing the details of the encoder layers, let us present a general formulation of multi-head attention, which is extensively employed in the proposed architecture. An attention function $A$ receives three matrices $\QQ \in \Re^{N_q \times d_k}$ (query), $\KK \in \Re^{N_k \times d_k}$ (key) and $\VV \in \Re^{N_k \times d_v}$ (value), and returns a matrix $\OO \in \Re^{N_q \times d_v}$, where each row is computed as a linear combination of rows from $\VV$, weighted by normalized dot-product similarity between rows of $\QQ$ and $\KK$, as follows:
\begin{equation}
\label{eq:attention}
A(\QQ, \KK, \VV) = \text{softmax}\left( \frac{\QQ \KK^\top}{\sqrt{d_k}}\right) \VV
\end{equation}
where softmax is applied on rows of the input matrix.
In multi-head attention, in order to capture several possible attention patterns between elements, $\QQ$, $\KK$ and $\VV$ are usually computed by linearly projecting a set of input matrices $\hat{\QQ} \in \Re^{N_q \times \hat{d}_k}$, $\hat{\KK} \in \Re^{N_k \times \hat{d}_k}$ and $\hat{\VV} \in \Re^{N_k \times \hat{d}_v}$ through multiple sets of projection matrices $\left\{ \left( \WW_{q,i}, \WW_{k,i}, \WW_{v,i} \right) \right\}_{i=1,\dots,h}$, with $h$ being the number of heads. The attention outputs for each set of projection matrices are then concatenated and linearly projected to produce the final output, as follows:
\begin{equation}
\text{MA}(\hat{\QQ}, \hat{\KK}, \hat{\VV}) =
\text{concat}\left(H_1, \dots, H_h\right) \WW_O
\end{equation}
where $\WW^O \in \Re^{N \times d_o}$ is a linear projector to the desired output dimension, and $H_i$ is the output of the $i$-th attention head:
\begin{equation}
H_i = A\left( \hat{\QQ}\WW_{q,i}, \hat{\KK}\WW_{k,i}, \hat{\VV}\WW_{v,i} \right)
\end{equation}
The amount of computation required for multi-head attention is approximately the same as in single-head attention, by uniformly splitting dimensions $d_q$, $d_k$ and $d_v$ among the $h$ heads.
In this work, for simplicity, we set $d_q = d_k = d_v = d_o = d$, whose specific value depends on where multi-head attention is employed in the network, as described below.

\begin{figure}
\begin{center}
\includegraphics[width=0.8\linewidth]{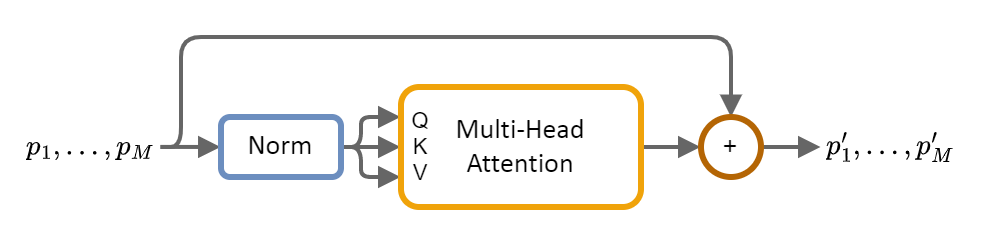}
\end{center}
   \caption{Architecture of the self-attention module for cluster tokens.}
\label{fig:sa_cluster}
\end{figure}

\noindent \textbf{Self-attention for cluster tokens.} The architecture of the self-attention module for cluster tokens is presented in Fig.~\ref{fig:sa_cluster}. The module receives the set of cluster tokens $\PP$ and applies a function $\text{SA}_p$ defined as:
\begin{equation}
\PP_n = \text{PLN} \left( \PP \right)
\end{equation}
\begin{equation}
\text{SA}_p \left( \PP \right) = \text{MA}\left( \PP_n, \PP_n, \PP_n \right) + \PP
\end{equation}
where $\text{PLN}$ is Pre-Layer Normalization~\cite{xiong2020layer}, which has been shown to improve training of transformer architectures, and query, key and values matrices are all set to $\PP_n$, as is typical of self-attention. A final residual connection is applied to improve gradient flow. The $d$ size is set to $d_p$, i.e., the size of the input cluster tokens.

\begin{figure}
\begin{center}
\includegraphics[width=0.8\linewidth]{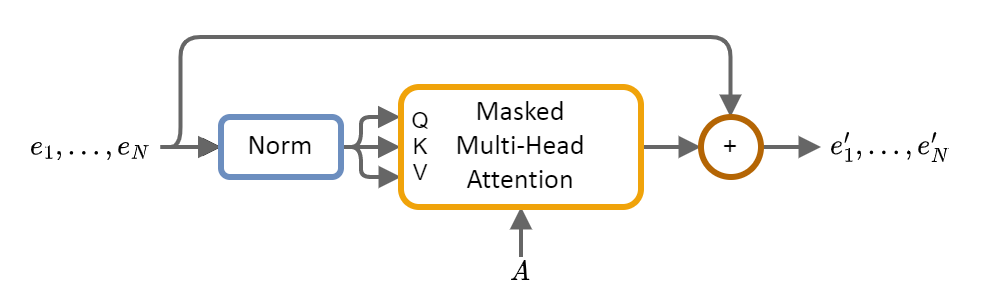}
\end{center}
   \caption{Representation of the self-attention module for triangles.}
\label{fig:sa_triangles}
\end{figure}

\noindent \textbf{Self-attention for triangles tokens}, illustrated in Fig.~\ref{fig:sa_triangles}, shares the same architecture as the self-attention module for clusters, but it employs the adjacency matrix $\AA$ as a mask for multi-head attention computation. The choice to adopt a adjacency-based attention masking mechanism is due to the need to preserve the capacity of the model to capture both local composition and long-range dependency~\cite{guo2019star} and to reduce computation requirements for high-resolution meshes exploiting the sparsity of the $\AA$ matrix. To carry out masked multi-head attention, the attention function in Eq.~\ref{eq:attention} is modified by subtracting infinity from masked positions of the query-key similarity vector, in order to nullify the corresponding softmax terms. The resulting attention function $A_\text{mask}$ is defined as:
\begin{equation}
A_\text{mask}(\QQ, \KK, \VV, \MM) = \text{softmax}\left( \frac{\QQ \KK^\top - \MM}{\sqrt{d_k}}\right) \VV
\end{equation}
where elements of $\MM$ are either 0 or $-\infty$. We can thus define our self-attention module for triangles as: 
\begin{equation}
\EE_n = \text{PLN} \left( \EE \right)
\end{equation}
\begin{equation}
\text{SA}_t \left( \EE \right) = \text{MA}_\text{mask}\left( \EE_n, \EE_n, \EE_n, \hat{\AA} \right) + \EE
\end{equation}
where $\text{MA}_\text{mask}$ is the variant $\text{MA}$ employing $A_\text{mask}$ as attention function, and $\hat{\AA} = \log\AA$, so that $\hat{A}_{ij} = -\infty$ where $A_{ij}$ = 0, and $\hat{A}_{ij} = 0$ where $A_{ij}$ = 1. The $d$ size for multi-head attention is set to $d_t$, i.e., the size of the input triangle tokens.

\begin{figure}
\begin{center}
\includegraphics[width=0.8\linewidth]{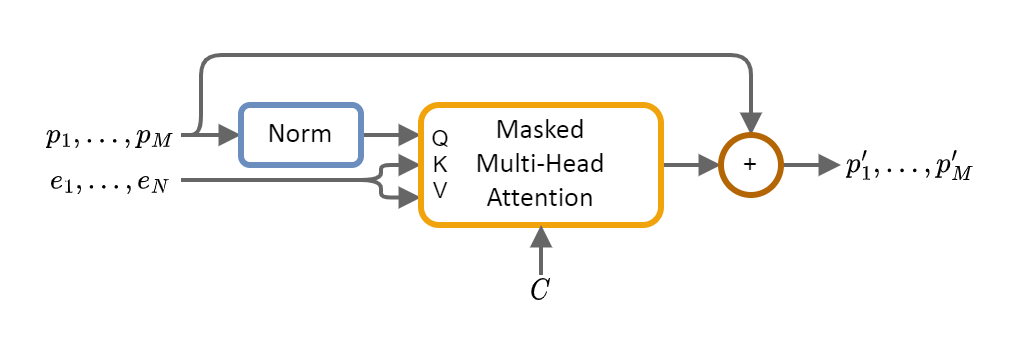}
\end{center}
   \caption{Representation of the cluster-triangle update module.}
\label{fig:ca_cluster}
\end{figure}

\noindent \textbf{Updating cluster representation from triangle tokens}. The cluster-triangle update module is introduced to
update the clusters' representation w.r.t. the triangles', thus allowing the network to exchange information between the two different modalities employed for modeling graph structure, i.e., Laplacian eigenvectors and clustering. To this aim, we employ masked multi-head attention using cluster tokens for computing query vectors, and triangle tokens to compute keys and values; in order to aggregate, for each cluster, only information of the triangles contained in it, we compute a symmetric matrix $\CC$ from the $\JJ$ clustering matrix by setting $C_{ij} = 1$ if triangles $i$ and $j$ belong to the same cluster, i.e., $\jj_i = \jj_j$, and $C_{ij} = 0$ otherwise.
The architecture of the cluster-triangle update module is presented in Fig.~\ref{fig:ca_cluster}, and implements the following function:
\begin{equation}
\PP_n = \text{PLN} \left( \PP \right)
\end{equation}
\begin{equation}
\text{CT} \left( \EE, \PP \right) = \text{MA}_\text{mask}\left( \PP_n, \EE, \EE, \hat{\CC} \right) + \PP
\end{equation}
where mask $\hat{\CC} = \log\CC$, as above. The $d$ dimension for multi-head attention is set to $d_p$, i.e., the size of input cluster tokens.

\begin{figure}
\begin{center}
\includegraphics[width=0.6\linewidth]{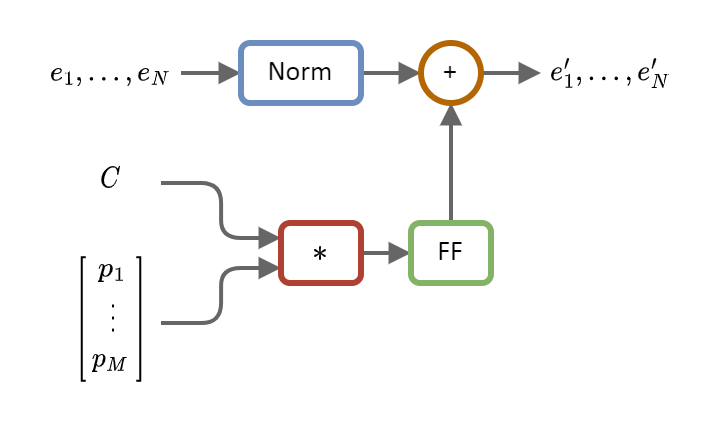}
\end{center}
   \caption{Representation of the triangle-cluster update module.}
\label{fig:ca_triangles}
\end{figure}

\noindent \textbf{Updating triangle representations from cluster tokens}.
A triangle-cluster update module is also used to update triangle representation with respect to clusters. Similarly to the cluster-triangle case, each triangle is affected only by elements belonging to the same cluster. The cross-attention module computes the sum between each triangle token and a projection of the average of the corresponding cluster tokens through a single feed-forward layer, as follows:
\begin{equation}
\EE_n = \text{PLN} \left( \EE \right)
\end{equation}
\begin{equation}
\text{TC}\left( \EE, \PP \right) = \EE_n + \text{FF}_\text{TC}\left( \CC\PP \right)
\end{equation}
where $\text{FF}_\text{TC}$ is a single feedforward layer.
The architecture of the triangle update module is presented in Fig.~\ref{fig:ca_triangles}. This operation can be interpreted as a form of cross-attention between triangle tokens and cluster tokens, where the former attend to the latter by means of a constant attention factor defined by cluster membership.

\noindent\textbf{Layer residual connection}.
The output of each token stream of a network layer finally undergoes a feedforward residual transformation, to independently transform each token, as follows:
\begin{equation}
\SS_n = \text{PLN} \left( \SS \right)
\end{equation}
\begin{equation}
\left( \SS \right) = \text{FF}_R\left( \SS_n \right) + \SS
\end{equation}
where $\SS$ is either $\EE$ or $\PP$, and $\text{FF}_R$ is a feedforward layer.
The architecture of the triangle update module is presented in Fig.~\ref{fig:ca_triangles}.
\section{Experimental results}
\label{sec:results}

\noindent In this section, we first introduce the datasets employed in our work: the COSEG Shapes dataset~\cite{wang2012active}, and the Human segmentation datasets proposed by~\cite{maron2017convolutional}.
Then, we evaluate the accuracy of our approach on the two different datasets. First, we assess how the model performs in three categories of the COSEG dataset, namely, \emph{Chairs}, \emph{Vases} and \emph{Tele-Aliens}; afterwards, we evaluate our method on the segmentation of human body meshes as well as on the ShapeNet dataset~\cite{chang2015shapenet}.
Ablation study then follows to substantiate the choices on the architecture components.
As a methodical note on the evaluation, for the comparison to \modelname,  with existing methods, we report the performance values reported in their original papers on the considered benchmarks. 

\subsection{Datasets and metrics}
\label{sec:satasets}
We test the performance of \modelname and compare it with those yielded by existing models on three standard benchmarks, namely, the Shape COSEG dataset~\cite{wang2012active}, the Human dataset~\cite{maron2017convolutional} and the ShapeNet dataset~\cite{chang2015shapenet}.\\
The Shape COSEG dataset~\cite{wang2012active}  consists of 11 sets of shapes with a consistent ground-truth segmentation and labeling: 8 sets are rather small and come from the dataset by~\cite{sidi2011unsupervised}, while the 3 remaining ones contain, respectively, tele-alines, vases and chairs. Given the scale of tele-alines, vases and chairs sets compared to the other eight sets, we used only them to evaluate the performance of \modelname. Train and test splits are the same defined in MeshCNN~\cite{hanocka2019meshcnn} for a fair comparison. As validation set we use  $6\%$ of the training set. \\
We also evaluate our method on human segmentation dataset introduced by~\cite{maron2017convolutional}. It consists of human meshes from several datasets, in particular SCAPE, FAUST, MIT Animation and SHREC 2007. The latest is used as test set, as in the MeshCNN~\cite{hanocka2019meshcnn} paper. \\
ShapeNet~\cite{chang2015shapenet} is a large-scale repository of shapes represented by 3D models of objects categorized following the WordNet taxonomy. ShapeNet contains semantic annotations about object parts as well as for rigid alignments, bilateral symmetry planes, physical sizes and other annotations.

\subsection{Model training and evaluation}

We train our model with mini-batch gradient descent, using the AdamW~\cite{loshchilov2017decoupled} optimizer and a batch size of 12. Learning rate is set to $5\cdot 10^{-5}$ with a weight decay of 0.01.
Dropout with probability 0.1 is used after each feedforward layer and multi-head attention in the transformer encoder, and after each feedforward layer in the classification network. 
The value of the $\lambda$ parameters controlling the features clustering, described in Sec.~\ref{sec:method}, is 8 for all the experiments, while token dimensions are set as $d_t = 512$ and $d_p = 1024$. All these parameters we set by measuring performance on a validation set extracted from each the COSEG dataset.
Cross-entropy loss function is used and weighted for each triangle based on its surface (larger triangles have more weight).
We perform data augmentation by applying random translation, rotation and scaling for each mesh in a mini-batch.\\
Accuracy is computed, as in DCN~\cite{xu2017directionally}, as the total surface of triangles correctly classified over the entire surface.

\noindent \textbf{Data preprocessing}. Similarly to MeshCNN, each mesh in the dataset is preprocessed reducing the number of vertices to a maximum of 1200 using the algorithm proposed by~\cite{garland1997surface}. Duplicated vertices are merged and ``padding'' triangles are added to allow batched processing of meshes. After preprocessing, each mesh consists of 2412 triangles. Padding triangles are not adjacent to any mesh triangle and do not influence the final prediction. Vertex coordinates are standardized between -1 and 1. The $\AA$ and $\PP$ matrices are extended to include the padding triangles.

\begin{figure*}
\begin{center}
\includegraphics[width=1\linewidth,height=11cm]{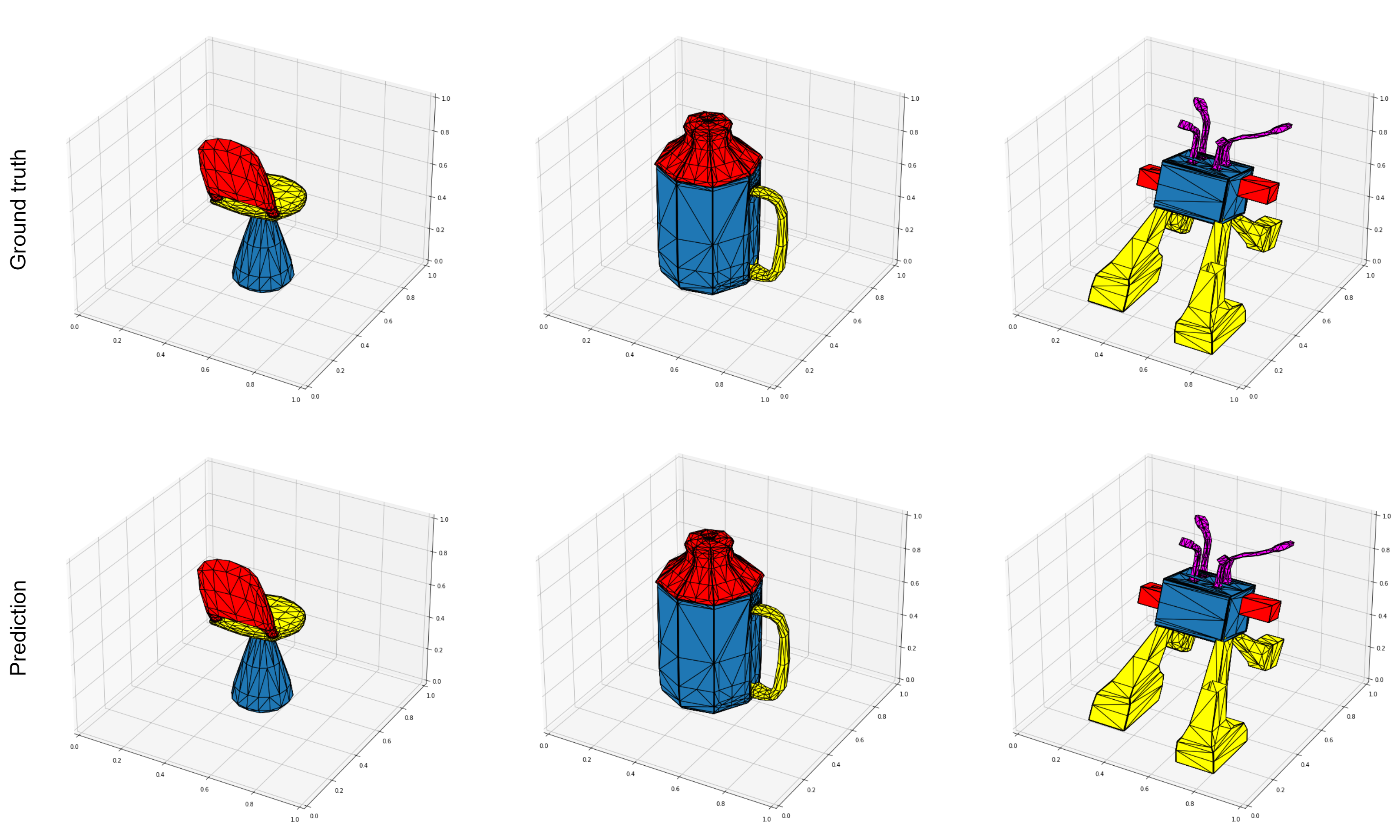}
\end{center}
\caption{Output segmentation examples by \modelname on COSEG meshes. At the top, the ground truth; at the bottom, the network prediction. Left to right: a chair example; a vase example and a tele-alien example.}
\label{fig:objects_sample}
\end{figure*}

We first evaluate our models on the Chairs, Vases and Tele-Aliens subsets of the COSEG dataset. For each set, we report the performance, in terms of accuracy.
Tab.~\ref{tab:coseg_results} shows that our approach achieves a higher global accuracy on all the COSEG sets w.r.t. state of the art methods. 
Fig.~\ref{fig:objects_sample} shows segmentation examples for each mesh set.

\begin{table}
\begin{center}
\begin{tabular}{l|c|c|c}
\textbf{Method}            & \textbf{Chairs} & \textbf{Vases} & \textbf{Tele-Aliens} \\
\hline
\hline
\cite{xie20143d} & 85.9 & 87.1 & 83.3 \\
\hline
\cite{kim2013learning} & 91.2 & 85.6 & -- \\
\hline
\cite{guo20153d} & 95.5 & 88.5 & -- \\
\hline
LaplacianNet \cite{qiao2019laplaciannet} & 94.2 & 92.2 & 93.9 \\
\hline
\cite{wang20183d} & 95.9 & 91.2 & 93.0 \\
\hline
\cite{xu2017directionally} & 95.7 & 90.9 & -- \\
\hline
MeshCNN \cite{hanocka2019meshcnn} & 99.6 & 97.3 & 97.6 \\
\hline
MeshWalker \cite{meshwalker} & 99.6 & 98.7& 99.1\\
\hline
NGD \cite{zhuang2022ngd} & 95.2 & 91.8 &  94.3 \\
\hline
MeshFormer \cite{li2022meshformer} & 98.9 & \textbf{99.1} & 99.1\\
\hline
\textbf{\modelname} (Ours) & \textbf{99.8} & 98.9 & \textbf{99.3} 
\end{tabular}
\end{center}
\caption{Quantitative results (classification accuracy in percentage) on the COSEG dataset. Values are reported in terms of accuracy between the predicted and ground-truth segmentation.}
\label{tab:coseg_results}
\end{table}

\begin{figure*}
\begin{center}
\includegraphics[width=0.9\linewidth,height=10cm]{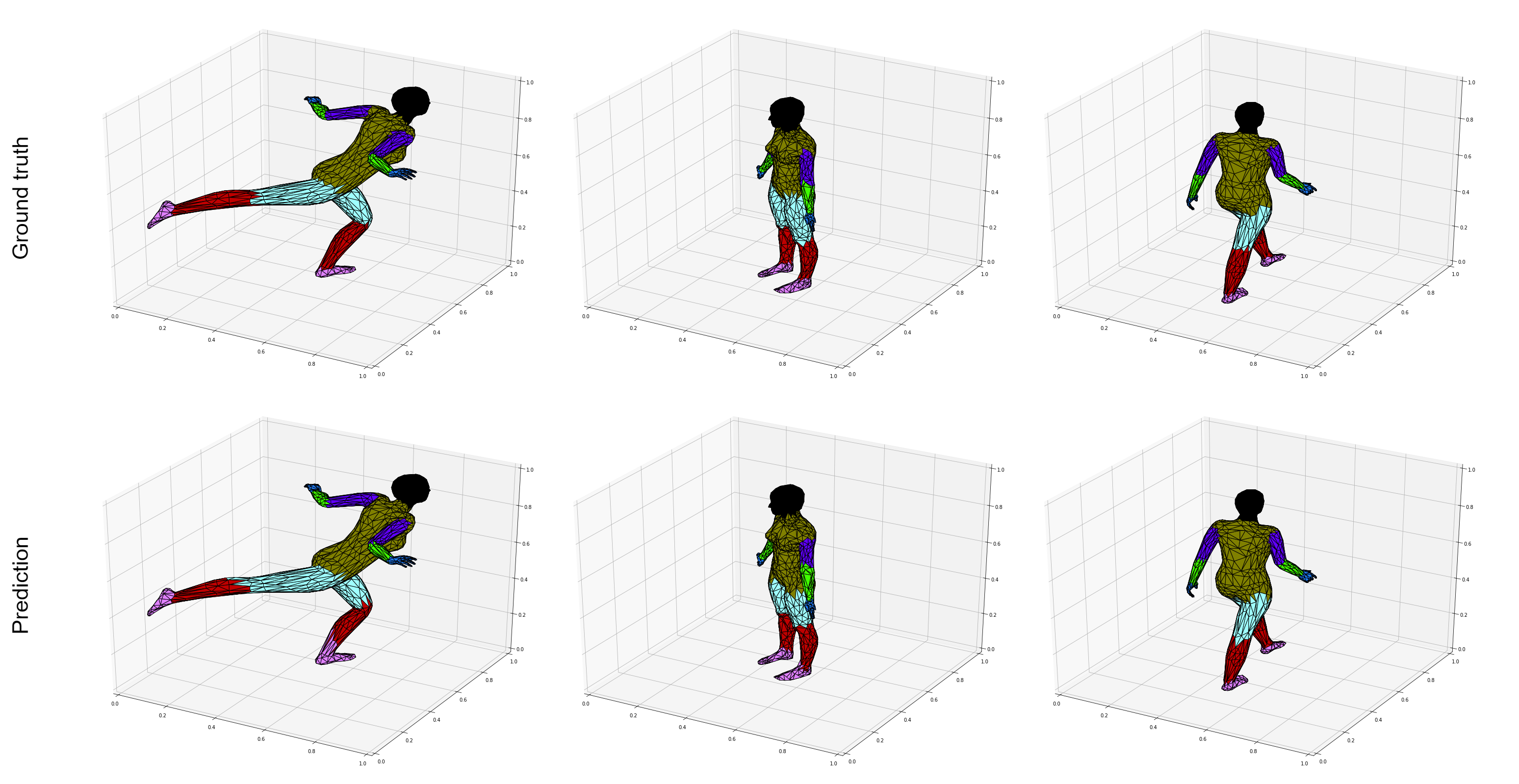}
\end{center}
\caption{Output segmentation samples on human meshes. At the top the segmentation ground truth; at the bottom the network prediction.}
\label{fig:humans_sample}
\end{figure*}

Mesh segmentation performances on the Human dataset~\cite{maron2017convolutional} are reported in Tab. \ref{tab:human_results}, showing better performance of our approach also on this benchmark when comparing with three state-of-the-art algorithms. Fig.~\ref{fig:humans_sample} shows qualitative results for the predicted segmentation. 

\begin{table}
\begin{center}
\begin{tabular}{l|c} \textbf{Method }& \shortstack{Laplacian encoding} \\
\hline
\hline
MeshCNN \cite{hanocka2019meshcnn}       & 92.3 \\
\hline
MeshWalker \cite{meshwalker}       & 92.7 \\
\hline
\cite{sharp:hal-03938034}       & 91.7 \\
\hline
\textbf{\modelname} (Ours) & \textbf{93.6} 
\end{tabular}
\end{center}
\caption{Quantitative results (classification accuracy in percentage) on the Human dataset. Values are reported in terms of accuracy between the predicted and ground-truth segmentation.}
\label{tab:human_results}
\end{table}

Finally, we compute mesh segmentation performances on the ShapeNet dataset~\cite{maron2017convolutional}, which are showed in Tab. \ref{tab:shapenet_results}. Also on this benchmark, \modelname yields better accuracy than state-of-the-art methods.

\begin{table}
\begin{center}
\begin{tabular}{l|c} \textbf{Method} & \textbf{ShapeNet}\\
\hline
\hline
Shapeboost \cite{Kalogerakis}       & 77.2 \\
\hline
\cite{guo20153d} & 77.6\\
\hline
ShapePFCN \cite{KalogerakisAMC17}       & 85.7 \\
\hline
LaplacianNet \cite{qiao2019laplaciannet} &  91.5\\
\hline
MeshTransformer \cite{li2022meshformer} &  92.6\\
\hline
\textbf{\modelname} (Ours) & \textbf{94.2} 
\end{tabular}
\end{center}
\caption{Quantitative results (classification accuracy in percentage) on the ShapeNet dataset. Values are reported in terms of accuracy between the predicted and ground-truth segmentation.}
\label{tab:shapenet_results}
\end{table}

\subsection{Ablation study}

We perform an ablation study, on the three subsets of the \emph{COSEG} dataset, to substantiate our design choices. 
We first assess how each component in the triangle representation affects performance, namely, triangle coordinates, surface normal and Laplacian positional encoding. Results in Tab.~\ref{tab:ablation_arch} show that all the input features positively affect accuracy. However, the highest contribution to the final performance is provided by the the Laplacian.

We then assess the importance of the cluster-related stream described in Sec.~\ref{sec:method}, i.e., cluster self-attention and cluster-triangle cross-attention . A comparison of the model accuracy with and without the cluster modules is presented in Tab. \ref{tab:ablation_arch_cluster}, where we can see that the cluster modules lead to gain of 3.6 percent points over the baseline, i.e., the model using only triangle information.

\begin{table}
\begin{center}
\begin{tabular}{l|c|c|c}
Model version & Chairs & Vases & Tele-Aliens \\
\hline
\hline
$w/o$ coordinates              & 94.4 & 92.1 & 93.8 \\
\hline
$w/o$ Laplacian embedding      & 97.9 & 97.0 & 97.4 \\
\hline
$w/o$ normals                  & 98.8 & 97.4 & 98.6 \\
\hline
\textbf{Full \modelname~input} & \textbf{99.8} & \textbf{98.9} & \textbf{99.3} 
\end{tabular}
\end{center}
\caption{\textbf{Ablation study of input features.} We ablate the transformer network on the input to the model.}
\label{tab:ablation_arch}
\end{table}

\begin{table}
\begin{center}
\begin{tabular}{l|c|c|c}
Model version& Chairs & Vases & Tele-Aliens \\
\hline
\hline
$w/o$ cluster modules     & 96.4 & 95.3 & 96.1 \\
\hline
\textbf{Full \modelname}  & \textbf{99.8} & \textbf{98.9} & \textbf{99.3}
\end{tabular}
\end{center}
\caption{\textbf{Ablation study of design choices.} We ablate the transformer network, comparing performance with and without the cluster modules.}
\label{tab:ablation_arch_cluster}
\end{table}

\section{Conclusion}
\label{sec:conclusion}
In this work, we introduce a novel transformer-based architecture for 3D mesh segmentation. Our approach successfully and significantly extends standard transformers with features specifically designed for the task at hand. First, we introduce a two-stream processing pipeline with each transformer layer, designed to enforce locality through the combination between mesh triangle features and clustering-based features, and by integrating spectral graph properties, through Laplacian vectors, to replace classic sinusoidal positional encoding. Additionally, we adapt typical attention mechanisms in transformers, by taking into account graph properties, and in particular by using adjacency matrix and triangle clustering to explicitly mask multi-head self- and cross-attention.\\
Experimental results, evaluated on multiple object categories, show that the resulting approach is able to outperform state-of-the-art methods on mesh segmentation, and demonstrate the positive impact of our architectural novelties by means of extended ablation studies. \\
To conclude, we show that transformer models --- in spite of their characteristics for global processing and limitations with representing locality in sparse graphs --- can be successfully adapted to mesh analysis, by carefully integrating methodological adjustments designed to capture mesh properties in a complex task such as segmentation.

\section{Acknowledgements}
This research is supported by the project Future Artificial Intelligence Research (FAIR) PNRR MUR Cod. PE0000013- CUP: E63C22001940006 and by the project “LEGO.AI: LEarning the Geometry of knOwledge in AI systems”, n. 2020TA3K9N.

{\small
\bibliographystyle{IEEEtran}
\bibliography{refs}
}

\end{document}